# A Simple and Efficient Probabilistic Language model for Code-Mixed Text

M Zeeshan Ansari[1]*, Tanvir Ahmad[1], M M Sufyan Beg[2], Asma Ikram[1]

[1] *Department of Computer Engineering, Jamia Millia Islamia, New Delhi, India*
[2] *Department of Computer Engineering, Aligarh Muslim University, Aligarh, India*

**Abstract**

The conventional natural language processing approaches are not accustomed to the social media text due to colloquial discourse and non-homogeneous characteristics. Significantly, the language identification in a multilingual document is ascertained to be a preceding subtask in several information extraction applications such as information retrieval, named entity recognition, relation extraction, etc. The problem is often more challenging in code-mixed documents wherein foreign languages words are drawn into base language while framing the text. The word embeddings are powerful language modeling tools for representation of text documents useful in obtaining similarity between words or documents. We present a simple probabilistic approach for building efficient word embedding for code-mixed text and exemplifying it over language identification of Hindi-English short test messages scrapped from Twitter. We examine its efficacy for the classification task using bidirectional LSTMs and SVMs and observe its improved scores over various existing code-mixed embeddings.

## 1. Introduction

The exponential growth in multifaceted data sources generated by social media platforms has led to the evolution of unusual complications in language processing. The speech and language processing task being considered a well-studied problem, multilingualism changed this notion and generated novel difficulties and challenges. Therefore, the prerequisite task in texts analysis of multiple languages is identification and segregation of the languages used in text [1]. A-priori language classification removes the language barrier in critical tasks involved in the social media text such as information extraction, sentiment analysis, and fake news detection. Most of the work until the last decade focused on identifying the languages present in mixed lingual documents, where each document belongs to a single language among the closed set of languages. Conversely, the recent studies primarily focus on code-mixed text language identification in which each document encompasses multiple languages within itself.

While the neural distributional language models are quite effective in capturing the meaning of words, Levy et al (2015)showed that traditional distributional models such as Pointwise Mutual Information (PMI) matrices produce comparable gains using specific system design preferences [2,3]. These embeddings are *d*-dimensional vectors of real numbers generated using word context pairs from a corpus. Despite that static embeddings are uncommon, we explore them due to the fact that, firstly, they do not require large volumes of data to train, specifically, in multilingual settings where data is limited, and secondly, they do not require much computing resources and get trained faster.

We perform an empirical evaluation of PMI based embeddings derived from code-mixed text. The proposed hypothesis in this work states that the code-mixed data encompasses strong *short span associations* between words of separate languages, for the extraction which, PMI, in principle, is remarkably suited. Consequently, simple and efficient embeddings may be generated for tasks related to code-mixed text, although complex embedding methods over monolingual text have proven valuable in natural language processing [4,5]. We chose to apply this hypothesis over Hindi-English code-mixed text obtained from Twitter, as a preliminary investigation. We identified users that posted an adequate mix of Romanized Hindi and English words and scrapped their tweets for this work. We train a classifier with proposed embeddings to predict the language and analyze the results. The paper is structured as follows; Section 2 comprises of related work Section 3 discusses theoretical dimensions to illustrate our approach towards the problem; Section 4 encompasses the overall methodology; Section 5 focuses on the experiments and obtained results; Section 6 summarizes the work in conclusion with future scope of work.

*Corresponding author
E-mail addresses: mzansari@jmi.ac.in,

## 2. Related Work

Traditional machine learning algorithms for automatic language identification extensively exploited the unique and consistent structures present in the words and sentences of a language. Baldwin and Lui (2010) utilized orthographic features and character-set mappings to represent a particular language feature set [11]. Word-level identification of languages explored by King and Abney (2013) [12] showed that performance improved by considering N-gram models coupled with Logistic Regression. Malmasi and Dras (2018) [8] used ensemble classifier-based language ldentification. Ansari et al. [9] inspected the effect of POS tags on learning of the language labels using the corpus of frequent English and transliterated Hindi words. Recent research works modeled language identification as a sequence labeling task using several deep neural networks, such as GRU, LSTM, and CNN. Jaech et al. (2016) [10] adopted Bidirectional LSTM to map the sequence of the CNN learned character vectors to the language label. Samih et al. (2016) [11] studied the effect of word and character embeddings with LSTM for Spanish-English language pairs from the Second Shared Task on Language Identification in Code-Switched Data.

Part of these development attributes to studies on different aspects of statistical and neural representations of word vectors. Jungmaier et al (2020) [12] applied Dirichlet smoothing on static embeddings to investigate its effect on rare words problem in low resource languages. In the works that show the influence of embedding matrix parameters over context vectors generated, the evaluation of corpus and document level properties affecting the PMI in semantic relatedness tasks is studied. [13]. Word and character n-gram vectors with classifiers for Hindi-English Social Media text yield superior results than word embeddings when adequate data for the corpus is available [ 14,15]. Shekhar et al. (2020) [16] applied quantum LSTM to detect language discriminations by utilizing bilingual lexicons. Language identification on several code-mixed Indian Languages is extensively studies in the shared task on Mixed Script Information Retrieval 2015 [17].

## 3. Background

### 3.1 Distributional language models

The notion of distributional language models characterizes a word as a vector in some multidimensional semantic space. Such vectors representing words, called embeddings, are easily learned from the corpus without any supervision. Similar words have similar vectors in semantic embedding space because they tend to occur in a similar context. The context is considered as a set of surrounding words of the target word in a window of fixed size. The definitions of context are extensively explored by Levy and Goldberg (2014) [18]. In contrast with document vector, word's vector representation is essentially a $|V| \times |V|$ word-context matrix of dimension equal to corpus vocabulary. Each element is the measure for association of a word with its context. Note that $|V|$ is the size each word vector, and is generally the size of vocabulary.

### 3.2 Pointwise mutual information matrix

The Pointwise Mutual Information (PMI) is an important metrics often used to measure the mutual dependence between two variables. For a given corpus, word *w* and its context *c* is associated using *PMI* as (Church and Hanks, 1989) [19]

$$PMI(w,c) = \log_2 \frac{P(w,c)}{P(w)P(c)} \qquad (1)$$

The *Positive PMI* matrix can be calculated for a vocabulary set using the word context matrix for $i^{th}$ word, $w_i$ and $j^{th}$ context, $c_j$, which is obtained as

$$PPMI(w_i, c_j) = \max\left(\log_2\left[\frac{p(w_i, c_j)}{p(w_i)p(c_j)}\right], 0\right) \qquad (2)$$

where, $p(w_i,c_j)$, $p(w_i)$ and $p(c_j)$ may be obtained using maximum likelihood estimation.

### 3.2 Singular value decomposition

A matrix *M* is factorized using truncated Singular Value Decomposition (TSVD) which decomposes it into three matrices as

$$M = U \cdot \Sigma \cdot V^T \qquad (3)$$

where *U* and *V* are in orthonormal form and Σ is a diagonal matrix of singular values in decreasing order. Reducing the representations of *M* to *d* dimensions as $M_d = U_d \cdot \Sigma_d \cdot V_d^T$, by retaining the top-*d* dimensions of Σ. The word embeddings can be obtained as $E = U_d \cdot \Sigma_d$ which are low dimensional representations for *PPMI* matrix on eq. (2).

## 4. Methodology

*4.1 Dataset preprocessing*

Dataset is prepared by searching for suitable handles on Twitter that contain a large amount of Hindi-English code-mixed text, and, subsequently, mining the tweets in such handles. The retrieved tweets belong to politics, sports, news, etc. The raw tweets extracted in the process contain noisy text viz., url, retweets, hashtags, emoticons, timestamps, unidentified special characters, etc., is thoroughly preprocessed. After the elimination of duplicate tweets, trailing spaces and newlines are also removed. The keyword 'url' replaces those tokens which do not belong to any language such as web links. Using usernames is a standard method in Twitter to direct specific tweets to particular users, those usernames replaced by 'user'.

*4.2 Annotation*

Keeping intact the tweets that contained all English and all Hindi words, for annotation of tokens, we used the tagset of seven entity classes: English (*En*), Hindi (*Hi*), Universal (*Univ*), Username (*User*), Hashtag (*Hash*), url (*U*) and Named Entity (*NE*). The annotation policies are relatively consistent with the other datasets derived from twitter [20,21]. We observed that, the usernames which are most frequent tokens follow very irregular orthographic patterns, due to which considering these would be trivial, moreover, identifying them would be comparable to Twitter Handle Classification Task which is not covered in this work. Subsequently, we preprocess the dataset through the normalization of all the usernames, along with the hashtags and urls by replacing their tokens with their respective labels. It is observed that there are several named entities, therefore, marked *NE*. The remaining words are annotated according to their natural language into *Hi* and *En* tags. The punctuations, emoticons, and rest of the other tokens which do not fall under the above six categories are marked as *Univ*. The word-level annotation is carried out by four annotators who are capable of understanding both Hindi and English language. The distribution of tags is shown in Table 1.

*4.3 Code Mixing Index*

Code Mixing Index (CMI) of Das and Gamback (2014) [22] is significant metrics to estimate the level of mixing between the languages while comparing different corpora. The CMI tends to give higher values when the words of both languages are high in numbers. On the other hand, if the number words of any language is low in number, the CMI tends to be low. We calculate CMI for bilingual settings of both Hindi and English language as

$$CMI = 1 - \frac{\max(N_{Hi}, N_{En})}{N_{Hi} + N_{En} + N_O} \quad (4)$$

where $N_{Hi} + N_{En} + N_O$ are the number of Hindi, English, and

Table 1

Distribution of tags in train and test samples.

| Tag | En | Hi | Univ | Hash | U | User | NE |
|---|---|---|---|---|---|---|---|
| #Train | 49467 | 20663 | 11719 | 3762 | 132 | 2756 | 3331 |
| #Test | 6908 | 1182 | 1413 | 212 | 21 | 210 | 254 |

tokens other than language respectively. Calculating the CMI for twitter corpus with 22.5% *Hi* tags and 58.8% *En* tags, yields the *CMI* value of 0.4475.

*4.4 Language model architecture*

The PMI gives the ratio of estimate for co-occurrence of two words together in a corpus, to their estimate of occurring together by chance. The positive values of PMI denote that the occurrence of two words together is higher than expected by chance of appearing them together and, negative values denote that their occurrence together is lower than expected by chance. However, high negative values may be produced by those words which occur less often due to which, a common technique applied is to use *Positive* PMI in which the negative values are replaced by 0.

From the data analysis of Hindi-English code-mixed text corpus, it is observed that the volume of Hindi vocabulary is far lower than English at sentence level as well as corpus level. This induces two properties (1) decrease in maximum likelihood estimates those Hindi words whose context is English and, (2) decrease in maximum likelihood estimates those English words whose context is Hindi. The maximum likelihood estimates required to calculate *PPMI* of eq. (1) are given according to

$$\left. \begin{array}{l} p(w_i, c_j) = \dfrac{f_{ij}}{\sum_{i=1}^{W} \sum_{j=1}^{C} f_{ij}} \\[6pt] p(w_i) = \dfrac{\sum_{j=1}^{C} f_{ij}}{\sum_{i=1}^{W} \sum_{j=1}^{C} f_{ij}} \\[6pt] p(c_j) = \dfrac{\sum_{i=1}^{W} f_{ij}}{\sum_{i=1}^{W} \sum_{j=1}^{C} f_{ij}} \end{array} \right\} \quad (5)$$

where, *W* and *C* denote the total number target words and their contexts respectively, which is generally equal to |V|.

Due to these two induced properties, the *PPMI* of bilingual word context pairs is maximized. Since the code-mixed text has abundance of such bilingual word pairs, the obtained *PPMI* matrix is having several high values. These high values are the strong associations captured between two separate language words in code-mixed text. This high dimension matrix is reduced to low dimensions using TSVD, and *PPMI* word embeddings are acquired. This low-rank approximation of *PPMI* matrix is used to generate the desired embeddings from train and test data, which is the input representation for the classifier.

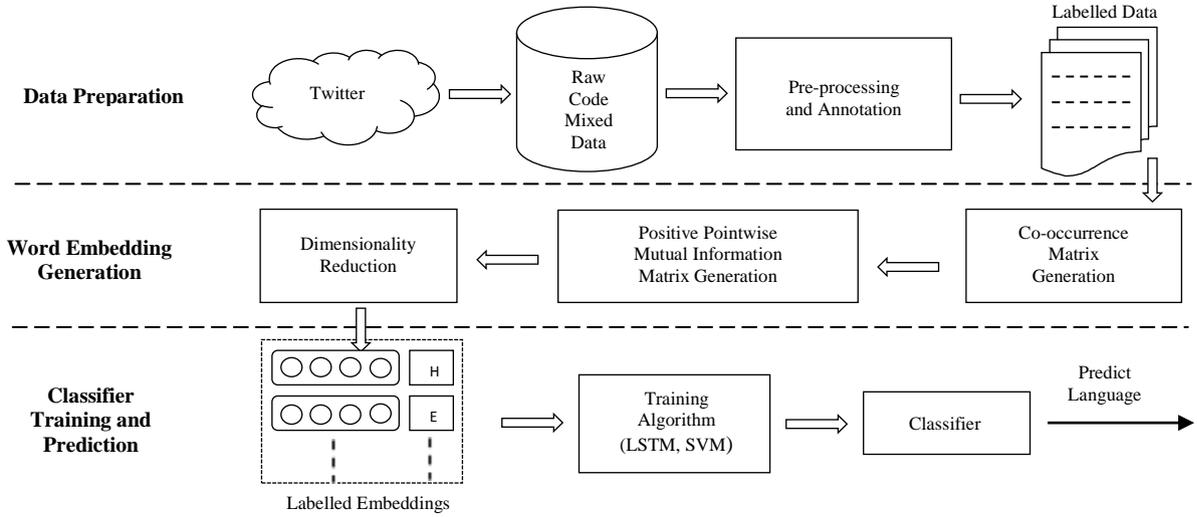

Fig. 1. Overall architecture

A bidirectional LSTM with softmax classifier is trained using default network parameters. An SVM classifier is also trained in order to assess the performance over a machine learning model in comparison to deep learning model. The overall proposed architecture is presented in Fig. 1.

## 5. Experiments and Result

The corpus of 5220 tweets is split with a 10:1 ratio into train and test data, and subsequently, 91800 train words and 10200 test words. The word co-occurrence matrix of shape $|V| \times |V|$ is obtained for the 4778 training tweets. Computing the *PPMI* from co-occurrence matrix and reducing it to $d$ dimensions, finally the embedding matrix of shape $|V| \times |d|$ is obtained, the value of $|V|$ is 15408 for our dataset. We consider $d$=100, 300, and 500 for our experiments. A look-up is performed in embedding matrix against each of the training words to produce 91800 embeddings of dimension $d$. Similarly, the look-up for test words results into 10200 embeddings of dimension $d$. It may be noted that, we have not handled out of vocabulary words while generating the embeddings.

With the aim to validate the proposed hypothesis, the extensive study is carried out over the PPMI values obtained for word context pairs present in code-mixed corpus. It was dramatically witnessed that values obtained for *bilingual* pairs are greater than monolingual pairs. For example, for the target word *media*, the PPMI value with Hindi context in *godi media* 8.87, however, on the other hand, the value with its English context is *paid media* is 5.84. Similarly, on computing the PPMI values for several word context pairs in respect of *media*, the bilingual pair average is 4.65 and its monolingual counterpart is 3.95. We carried out similar computations for other words, as shown in Table 2.

The classwise results of LSTM classifier with dimension $d$=100, as shown in Table 3, reveal that the classifier precision and recall of *Hi* tags are lower than *En* tags, whereas *NE* tags were most poorly classified with an F1-measure of 0.47 only. The detailed study of confusion between the classes discloses that the classifier rarely predicts any wrong *Univ* tags, however, approximately, 17% of *Univ* are predicted as *En*. After the inspection of such predicted tags, it was noted that mostly hyphenated English words were marked *Univ* during annotation. However, the *Hi* tags mispredicted as *En* tags could be studied using the statistical analysis of large size code-mixed corpus. It is also observed that the performance of classifier's improves marginally as we increase the dimension

Table 2

PPMI values of some word-context pairs and average of all pairs.

| target word = media | | | target word = election | | |
|---|---|---|---|---|---|
| word-context pair | code mixing | PPMI | word-context pair | code mixing | PPMI |
| *godi* media | yes | 8.87 | *sirf* politics | yes | 6.61 |
| *paid* media | no | 5.84 | *just* politics | no | 3.77 |
| media *ne* | yes | 2.41 | politics *aur* | yes | 3.23 |
| *by* media | no | 1.19 | politics *for* | no | 1.42 |
| all pairs average | yes | 4.65 | all pairs average | yes | 6.15 |
| all pairs average | no | 3.95 | all pairs average | no | 4.51 |

Table 3

Class-wise performance comparison of PPMI-LSTM

| Predicted Tag | P | R | F |
|---|---|---|---|
| English | 0.90 | 0.92 | 0.91 |
| Hindi | 0.64 | 0.71 | 0.67 |
| Named Entity | 0.48 | 0.46 | 0.47 |
| Universal | 0.98 | 0.80 | 0.88 |
| URL | 0.95 | 1.0 | 0.98 |
| Weighted Average | 0.88 | 0.87 | 0.87 |

Table 4

Comparison of F1-measure of proposed PPMI models with existing language models

| Model | English | Hindi | NE | Overall |
|---|---|---|---|---|
| PPMI-LSTM ($d=300$) | 0.913 | 0.674 | 0.475 | **0.871** |
| PPMI-SVM ($d=500$) | 0.905 | 0.643 | 0.480 | **0.865** |
| Veena et al (2017) [14] | 0.658 | 0.829 | 0.622 | 0.804 |
| Bhattu et al (2015) [15] | 0.831 | 0.613 | 0.387 | 0.769 |
| Sequiera et al (2015) [17] | 0.911 | 0.651 | 0.425 | 0.767 |
| Shekhar et al (2020) [16] | 0.857 | 0.939 | 0.839 | 0.742 |

size from 100 to 300 and 500.

The comparative analysis of proposed and existing models, in Table 4, shows that the performance of our approach is fairly better than other approaches and obtains highest average F1-measure. The PPMI-LSTM, when compared to PPMI-SVM performs better in terms of F1-measure. Moreover, in 30 epoch, it takes significantly less time with final training and validation losses equal to 0.1813 and 0.3367.

## 6. Conclusion and Future Scope

The authors have presented a simple and efficient approach for generating code-mixed vectors using Pointwise Mutual Information for Twitter Hindi-English text. Comparing the F1-measure of the recent existing works on word embedding generation approaches for Hindi-English code-mixed text, the *PMMI* based embedding surpasses them with substantial difference. The results can be certainly improved by correcting the shortcoming that were observed in the annotation of data through the error analysis. The potential improvement essential to the proposed approach is designing the parametric approach to construct the embedding matrix and use optimized parameters to train the classifier. Another significant improvement that may also be considered is the handling of out of vocabulary words, which will make the approach comparable to much recent language models based on fastText and BERT.